\documentclass[11pt,a4paper]{article}
\usepackage[dvipsnames]{xcolor}
\usepackage[hyperref]{acl2020}
\usepackage{times}
\usepackage{latexsym}

\usepackage{microtype}
\usepackage{xspace}
\usepackage{graphicx}
\usepackage{caption}
\usepackage{multirow}
\usepackage{multicol}
\usepackage{booktabs}
\usepackage{amssymb}
\usepackage{xspace}
\usepackage{enumitem}
\usepackage{comment}
\usepackage{fdsymbol}
\usepackage{etoolbox}

\usepackage{amsmath}

\usepackage[T1]{fontenc}

\usepackage[utf8]{inputenc}
\usepackage{contour}

\aclfinalcopy
\graphicspath{ {./graphics/} }

\definecolor{azure}{RGB}{0, 127, 255}
\definecolor{problematic}{RGB}{132, 94, 247}
\definecolor{figure_green}{RGB}{0, 176, 80}

\newcommand\jmhdontrender[1]{}

\newcommand\merlottitlefont[1]{{\color{Fuchsia} \textbf{{\smash{{\usefont{T1}{cmtt}{m}{n}#1}}}}}}
\newcommand\merlotfont[1]{\smash{{\usefont{T1}{cmtt}{m}{n}#1}}}
\newcommand{\modelnamewithlogo}{\logo\,\merlotfont{ESPER}\xspace}
\newcommand{\modelname}{\merlotfont{ESPER}\xspace}
\newcommand{\modelnamelong}{\merlottitlefont{E}xtra\merlottitlefont{S}ensory \merlottitlefont{PE}rception with \merlottitlefont{R}einforcement learning}
\newcommand{\datasetname}{ESP dataset\xspace}
\newcommand{\datasetlongname}{\merlottitlefont{E}valuation for \merlottitlefont{S}tyled \merlottitlefont{P}rompt dataset}

\newcommand{\logo}[0]{\text{\smash{\raisebox{-1pt}{\includegraphics[height=8pt]{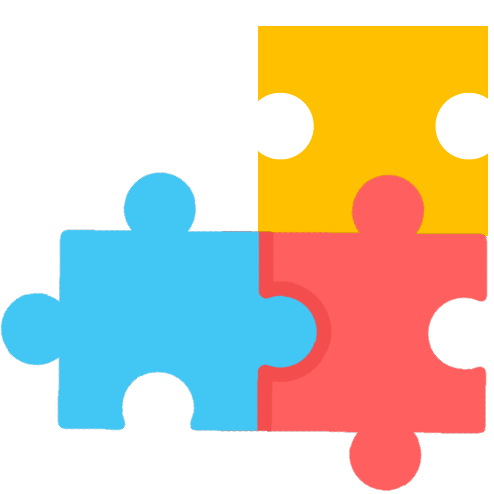}}}}}

\newcommand{\ie}{\textit{i}.\textit{e}., }
\newcommand{\eg}{\textit{e}.\textit{g}.\ }
\def\mathhyphen{{\hbox{-}}}

\newcommand*\samethanks[1][\value{footnote}]{\footnotemark[#1]}

\renewcommand{\spadesuit}[0]{\text{\smash{\raisebox{-1pt}{\includegraphics[height=6pt]{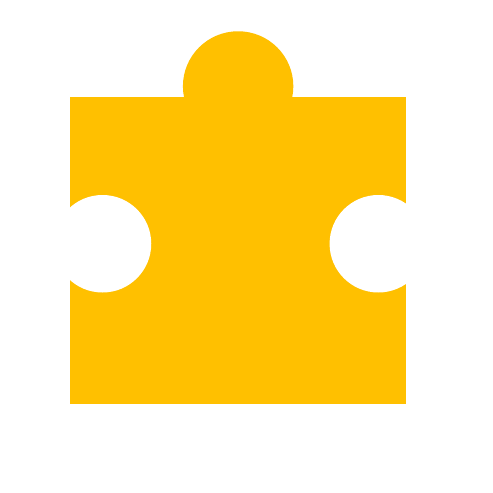}}}}}
\renewcommand{\heartsuit}[0]{\text{\smash{\raisebox{-1pt}{\includegraphics[height=6pt]{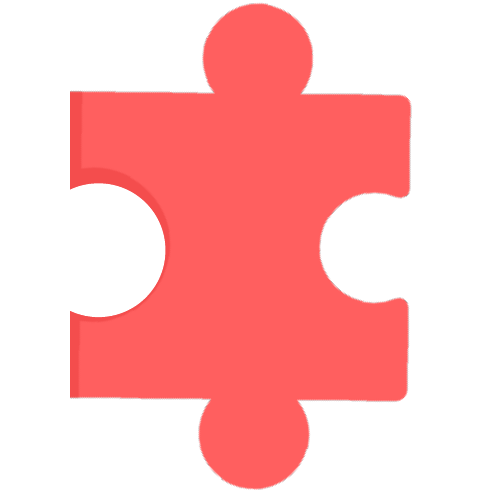}}}}}
\renewcommand{\clubsuit}[0]{\text{\smash{\raisebox{-1pt}{\includegraphics[height=6pt]{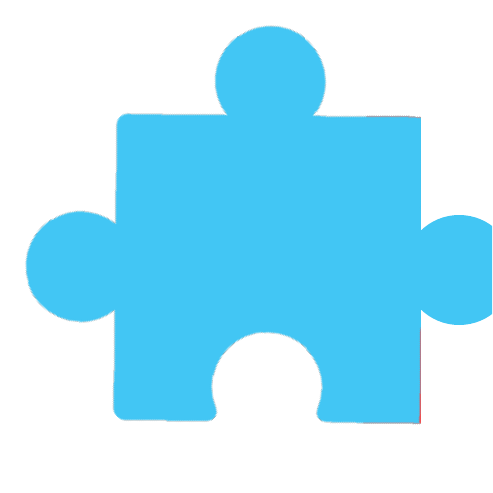}}}}}

\title{Multimodal Knowledge Alignment with Reinforcement Learning}

\author{
Youngjae Yu$^{\heartsuit}$\thanks{~~denotes equal contribution} \quad
Jiwan Chung$^{\spadesuit}$\samethanks \quad
Heeseung Yun$^{\spadesuit}$ \quad
Jack Hessel$^{\heartsuit}$ \quad \\
\textbf{Jae Sung Park}$^{\heartsuit\clubsuit}$ \quad
\textbf{Ximing Lu}$^{\heartsuit\clubsuit}$  \quad
\textbf{Rowan Zellers}$^{\clubsuit}$ \\ \quad
\textbf{Prithviraj Ammanabrolu}$^{\heartsuit}$ \quad
\textbf{Ronan Le Bras}$^{\heartsuit}$ \qquad
\textbf{Gunhee Kim}$^{\spadesuit}$ \qquad
\textbf{Yejin Choi}$^{\heartsuit\clubsuit}$ \qquad
\\
\normalsize{$\heartsuit$ Allen Institute for Artificial Intelligence}\\
\normalsize{$\spadesuit$ Department of Computer Science and Engineering, Seoul National University}\\
\normalsize{$\clubsuit$ Paul G. Allen School of Computer Science, University of Washington}\\
}

\date{}

\begin{document}
\maketitle
\begin{abstract}

Large language models readily adapt to novel settings, even without task-specific training data.
Can their zero-shot capacity be extended to \emph{multimodal} inputs?
In this work, we propose \modelnamewithlogo (\modelnamelong)
which extends language-only zero-shot models to unseen multimodal tasks, like image and audio captioning.
Our key novelty is to use reinforcement learning to align multimodal inputs to language
model generations without direct supervision:
for example, in the image case our reward optimization relies only on cosine similarity derived from CLIP \cite{radford2021learning}, and thus requires no additional explicitly paired (image, caption) data.
Because the parameters of the language model are left unchanged, the model maintains its capacity for zero-shot generalization.
Experiments demonstrate that \modelname outperforms baselines and prior work on a variety of zero-shot tasks; these include a new benchmark we collect and release, \datasetname, which tasks models with generating several diversely-styled captions for each image.

\end{abstract}
\section{Introduction}
\label{sec:intro}

Zero-shot learning challenges machine learning models to make inferences for novel tasks not explicitly seen at training time.
Recently, large, pretrained transformer-based models like GPT-3 \cite{brown2020gpt3} have achieved impressive zero-shot capabilities for a diverse set of language generation and reasoning tasks.
However, models like GPT-3 only accept textual prompts as input.

In this work, we propose a new model, \modelnamelong (\modelnamewithlogo), that
enables large language models
to accept multimodal inputs like images and perform broad generation tasks over those inputs.
In a zero-shot fashion, our model can
generate text diverse in style and context conditioned on an image, including visual news~\cite{liu2021visualnews}, visual dialogues~\cite{schwartz2021visdial}, answers to visual questions~\cite{antol-iccv-2015,goyal-cvpr-2017},
visual blog-style posts~\cite{kim2015joint}, and visual stories~\cite{huang-NAACL-2016}.

\begin{figure}[t]
\centering
\includegraphics[trim=0.0cm 0.0cm 0cm 0.0cm,clip,width=0.50\textwidth]{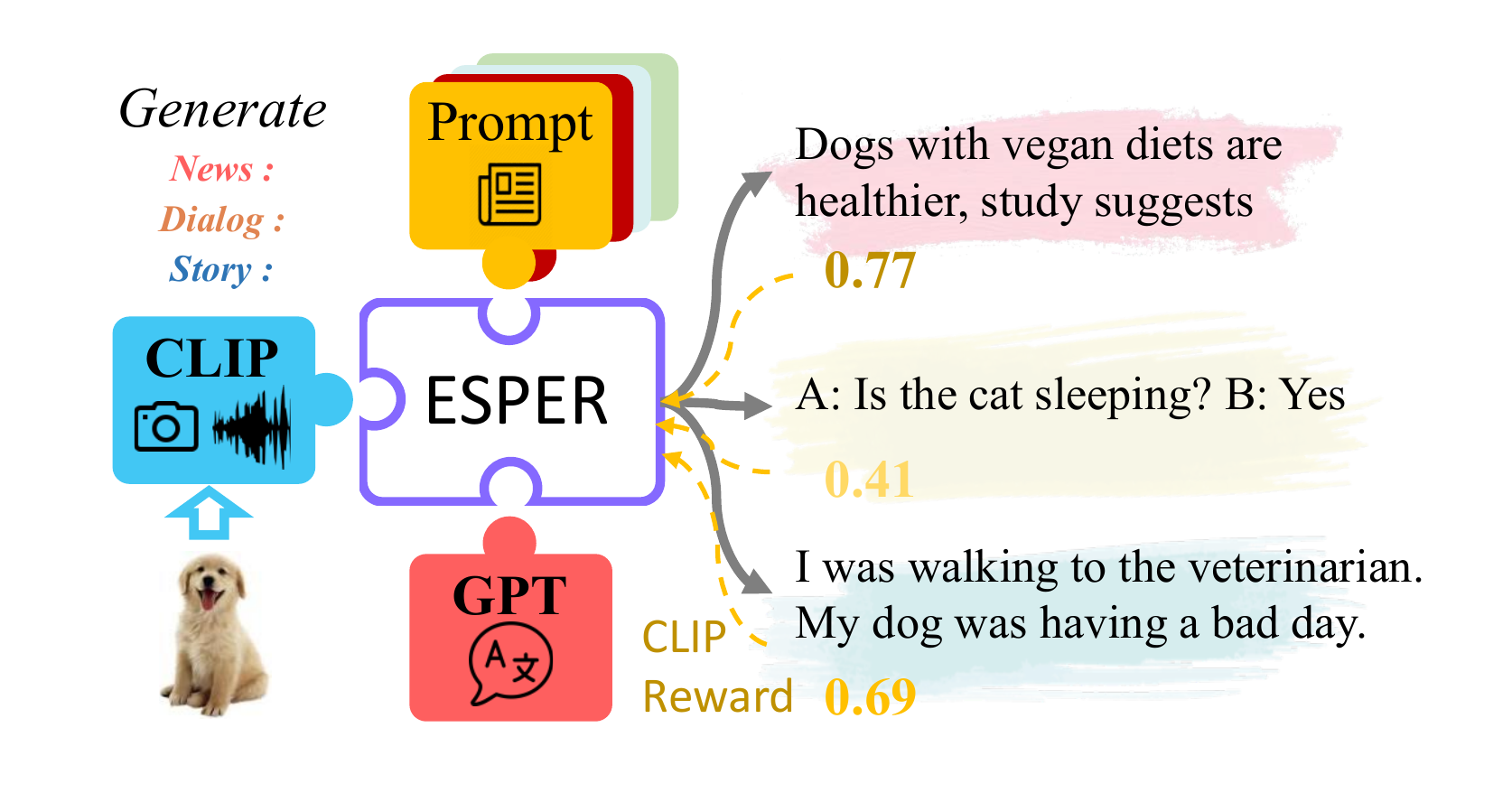}
\caption{The intuition of  \modelnamewithlogo, \modelnamelong.
  To better align knowledge in CLIP and GPT with RL,
  we give use CLIP for give rewards for pairs of images and self-generated text.
}

\label{fig:key_idea}
\end{figure}

\modelnamewithlogo achieves this by combining insights from two previously disjoint lines of work:
\emph{multimodal prompt tuning}, and \emph{reinforcement learning reward optimization}.
Like prior multimodal prompt tuning work, \modelname starts from a base language-only
model (e.g., GPT-2 \cite{radford2019language}), keeps most of its parameters frozen
and trains a small number of adaptor parameters to map visual features into the vocabulary space of the language model \cite{tsimpoukelli2021multimodal,mokady2021clipcap,liu2021gpt}. \emph{Unlike} prior works, however,
\modelname does not train these parameters using maximum likelihood estimation over a dataset
of aligned (image, caption) pairs. Instead, it uses a reinforcement learning objective.
During training, the model is first queried for completions conditioned on visual features.
Then, parameters of a lightweight vision-to-text transformation are updated using proximal policy
optimization (PPO) \cite{schulman2017ppo} to maximize
a similarity score computed by a secondary pretrained image-caption model, CLIP \cite{radford2021learning}.
The frozen language model can interpret the multimodal inputs in the same context
as the initial word embedding space without additional human-annotated paired data. 

A key advantage of using a reinforcement learning objective instead of a maximum likelihood
objective is the maintenance of generalizability.
\newcite{tsimpoukelli2021multimodal,mokady2021clipcap} fine-tune their lightweight visual-to-language adapters using paired visual-linguistic datasets such as Conceptual Captions~\cite{sharma2018conceptual} or COCO Captions~\cite{lin2014microsoft}.
Because these datasets of literal descriptions cannot match the textual variety of
the large-scale corpus GPT-2 is trained on,
the supervised models may not generate as richly styled language or be capable of as diverse reasoning over input contexts ~\cite{kumar2022finetuning,wortsman2022model}. %

We experimentally compare \modelname to two classes of prior methods that seek to adapt language models to accept visual inputs: (1) maximum likelihood prompt tuning \cite{tsimpoukelli2021multimodal,mokady2021clipcap}; and (2) decoding-time methods~\cite{tewel2022cvpr} that post-process token probabilities of a frozen language model according to estimated image similarity. For zero-shot image/audio captioning, we find that 
\modelname outperforms all prior unsupervised methods,
both in terms of generation quality (e.g., $14.6$ point improvement in CIDEr over~\newcite{laina2019iccv} in COCO unpaired captioning)
and inference speed (e.g., $10^2 \times$ speedup vs \newcite{tewel2022cvpr}, which relies on per-token gradient optimization over partial decodings.) %

In addition: 
(1) \modelname exhibits strong zero-shot adaptability on visual news~\cite{liu2021visualnews}, visual dialogue dialogue~\cite{das2017visual}, and a new zero-shot multimodal generation benchmark we construct+release called \datasetname,
which tests model capacity to generate texts of different styles for the \emph{same} image;
(2) we show that \modelname can learn about audio inputs using an audio-based reward.
We hope the strong performance of \modelname presented here will encourage researchers to consider RL-based training for future multimodal prompt tuning work, e.g., as a complement to max likelihood
models like Flamingo-80B \cite{alayrac2022flamingo}.

\section{Method}
\label{sec:method}

\begin{figure*}[t!]
	\begin{center}
		\includegraphics[trim=0.0cm 0.0cm 0cm 0.0cm,width=0.9\textwidth]{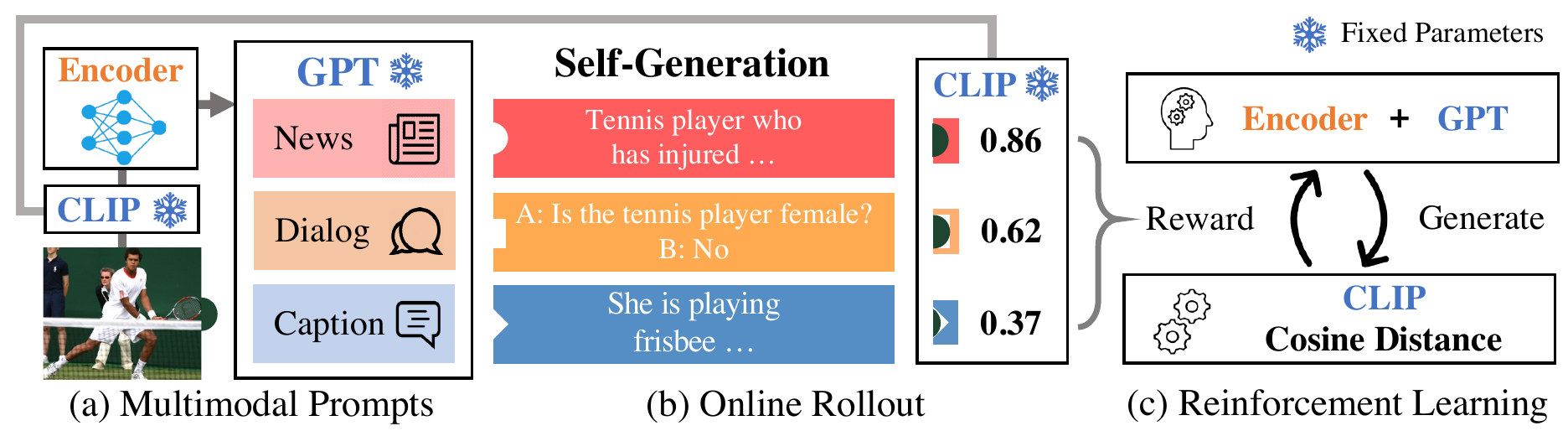}
	\end{center}
	\caption{Illustration of the proposed model, \modelname. We use a pretrained language model (\eg GPT-2~\cite{radford2019language}) as the language generator
}
	\label{fig:architecture}
\end{figure*}
 
\modelnamewithlogo consists of three components: 1) CLIP's non-generative image/text encoders \cite{radford2021learning};\footnote{While we describe image modeling here, we also experiment with audio/text encoders, specifically Wav2CLIP \cite{wu2022wav2clip}, in \S~\ref{subsec:audio_captioning} that extend \modelname to audio inputs.}
2) GPT-2 \cite{radford2019language}, a left-to-right language generator; and 3) an encoder that
projects multimodal inputs into the word embedding space of GPT-2.\footnote{In principle, any models with the same APIs could be used, e.g., ALIGN~\cite{Jia2021ScalingUV} could be substituted for CLIP, or T5~\cite{raffel2020t5} could be substituted for GPT-2 } %
During training, CLIP and GPT-2's parameters are frozen; gradients are back-propagated through the frozen language model to train the encoder parameters.
We employ reinforcement learning (specifically, PPO \cite{schulman2017ppo}) to derive these gradients: the reward function is the similarity of the sampled generations to the input image, as estimated by CLIP.
After RL training, we evaluate \modelname in various zero-shot scenarios.

\subsection{Architecture}
\label{sec:architecture}

\paragraph{CLIP.}
\newcite{radford2021learning}'s Contrastive Language Image Pretrained (CLIP) encoder plays two roles in our framework:
first, as a feature extractor for the input images, and second, as an alignment reward scorer between the images and the model-generated text.
First, the CLIP image encoder $CLIP\mathhyphen{}I$ extracts single vector feature from the image $x^i$.
Importantly, we do not update CLIP's parameters during training:
in practice, we extract features for all images prior to training for faster execution.
Second, the CLIP text encoder $CLIP\mathhyphen{}T$ is applied to text samples the model generates to support RL training; %
Combined with the pre-extracted image representation, this text representation is used to compute the reward function as the cosine similarity between the image and the model-generated text.
While CLIP's textual representations cannot be pre-cached like the image representations because the model's generations are dynamic, because we do not backpropagate gradients to the text network this process is fast and memory-efficient to run on a GPU.

\paragraph{Encoder.}
The encoder $F_{\phi}$ is the only module with trainable parameters in \modelname.
Given the vector representation of an image $x^i$ extracted using CLIP,
the module outputs a series of vectors of length $k$ to be passed on to the language model.
The output image representations $h^i$ work as the multimodal prompt and are concatenated to the embedded word representations.
We fix the visual token length in all experiments to $k=10$.
\begin{gather*}
h^i = h^i_1, \ldots, h^i_k = F_{\phi}(CLIP\mathhyphen{}I(x^i))
\end{gather*}
For fair comparison in later experiments, we use the same multimodal encoder architecture as CLIPCap~\cite{mokady2021clipcap}: a lightweight, two-layer Multi-Layer Perceptron (MLP).
The first layer maps the CLIP encoding dimensions to GPT-2's dimensions
and the second layer expands the single vector representation
to a series of vector representations of length $k$.
We use $tanh$ as the nonlinear activation function between these two layers.
By employing a less expressive encoder architecture (than, e.g., a transformer),
we aim to demonstrate that the contribution of \modelname
does not rely on the structure/capacity of the encoder itself.

\paragraph{Pretrained Language Model.}
\modelname employs a pretrained deep autoregressive language model such as GPT-2 \cite{radford2019language}
as the backbone.
Autoregressive language models parameterize likelihood of a text sequence
$y$ comprised of text tokens $y_j$ with length $l$ using autoregressive decomposition.
\begin{gather*}
p_{\theta} (y) = \prod_{j}^{l} p_{\theta} (y_j | y_{j' < j}) 
\end{gather*}
Inspired by prompt tuning in the text-only domain~\cite{liu2021gpt},
we treat the encoded image vector sequence $h^i$ as a multimodal prompt
and concatenate it with the text prompt representation
output by GPT-2's embedding lookup layer given previous tokens $y^i_{j' < j}$
to build the prefix for the conditioned text generation:
\begin{gather*}
p_{\theta} (y^i|h^i) = \prod_{j}^{l} p_{\theta} (y^i_j | h^i,  y^i_{j'}) 
\end{gather*}
The text prompt $z$ can be as short as a single %
word token for free-form training
or contain task-specific templates for further zero-shot adaption to downstream tasks.

The parameters of the language model $\theta$ are %
kept frozen.
However, the encoder parameters $\phi$ are updated with the gradients calculated
based on the language model parameters.
Hence, we connect multimodal information to the language model
without modifying the linguistic knowledge stored in the pretrained weights.

\begin{figure*}[t]
\centering
\includegraphics[trim=0.0cm 0.1cm 0cm 0.0cm,clip,width=0.95\textwidth]{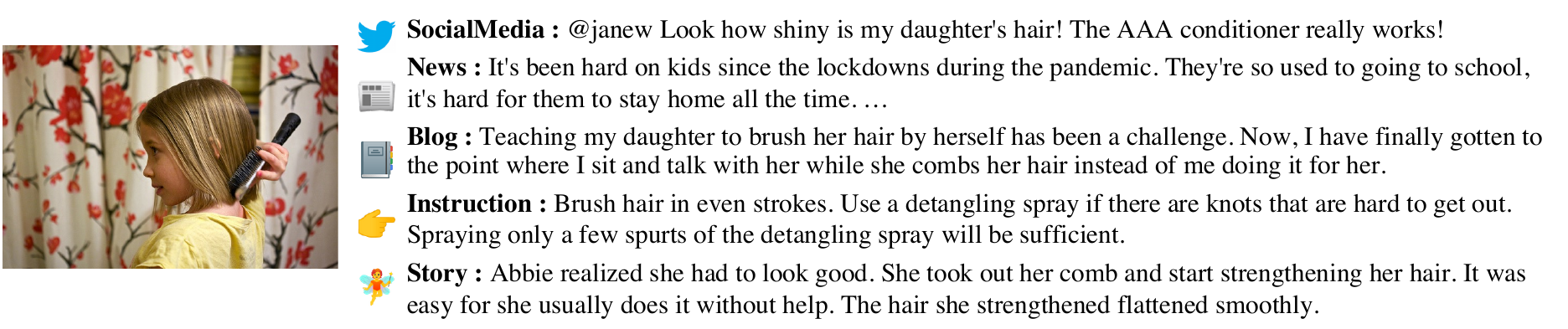}
\caption{A sample in \datasetlongname \; (\datasetname). 
}
\vspace{-7pt}
\label{fig:espdata}
\end{figure*}

\subsection{Training}
\label{sec:training}

\paragraph{Reinforcement Learning.}
Because CLIP %
does not provide per-token feedback, %
there is no directly differentiable way to train the encoder parameters to generate captions that CLIP would score highly, given the input image. %
Thus, we propose to view CLIP as a black-box model and apply reinforcement learning to minimize the embedding distance between the image context and the corresponding generated text. %
We use the clipped version of Proximal Policy Optimization (PPO-clip)~\cite{schulman2017ppo,stiennon2020learning} for reward optimization. 
From the RL perspective, our GPT-2 generator can be viewed as a policy, which produces actions (in the form of generations) given states (in the form of text+image prompts).
Our value model has the same architecture as \modelname; we use random sampling with temperature $0.7$ for text generation during training.

\paragraph{Modality Pairing Reward.}
The primary objective of \modelname is to align multimodal inputs to text generations.
Given an input image $x$ and the corresponding generated text $y$,
we regard the cosine similarity between the respective CLIP features
as the pairing reward.
\begin{gather*}
r^p(x, y) \approx \frac{CLIP\mathhyphen{}I(x)}{||CLIP\mathhyphen{}I(x)||} \cdot \frac{CLIP\mathhyphen{}T(y)}{||CLIP\mathhyphen{}T(y)||}
\end{gather*}
The actual reward is further normalized to roughly achieve zero mean and unit variance
over the course of training.
In practice, we multiply the cosine similarity value with a fixed gain ($\alpha=50$) and then add a fixed bias ($\beta=-10)$.

\paragraph{Language Model Stability.}
Reward hacking can potentially occur ~\cite{krakovna2020specification} if the
agent discovers incoherent texts that nonetheless achieve high rewards.
To defend against this, we incorporate a set of auxiliary rewards to stabilize the training process. 
First, we compute the KL divergence between $p_\theta$ and a separate (fixed) text-only GPT-2 model to maintain language generation capability.
In addition, we found it beneficial to consider raw text-only likelihood as an additional reward.
Finally, as reported in previous literature~\cite{holtzman2020topp,welleck2019unlikelihood}, language models tend to
falsely assign high probability on repetitive phrases.
We introduce an explicit repetition penalty against this phenomenon.
For specifics on the collection of stability rewards we apply, we refer interested readers to Appendix~\ref{sec:ax_stability_rewards}.

\subsection{\modelname-Style}
\label{sec:style_ptuning}
Following previous literature on adapting language models using prompts~\cite{gao2021making},
we consider a version of \modelname, where we pre-fine-tune GPT-2 with a text-only corpus alongside corresponding style prompt prefixes (\ie \texttt{"news:"}, \texttt{"story:"}).
For instance, to train a news generator we present the model with a news corpus~\cite{liu2021visualnews} 
prefixed with the style prompt (\texttt{"news:"}).
In practice, we finetune a single GPT on multiple styles.
Note that style prompt training uses only text corpus and does not require multimodal inputs.
We train these style-augmented GPT-2 generators prior to applying \modelname and provide them as backbones in place of the unconditional language models. %

\section{Experiments}
\label{sec:experiments}

\paragraph{\datasetname.}
\label{sec:data}
To benchmark \modelnamewithlogo's capability to generate diverse styles of writing from the \emph{same} image, %
we collect a novel dataset: \datasetname (\datasetlongname).
\datasetname is a benchmark for zero-shot diverse caption generation.
It comprises 4.8k captions from 1k images in the COCO Captions test set~\cite{lin2014microsoft}.
We collect five different writing styles that are frequently used, namely blog, social media, instruction, story, and news, as illustrated in Figure~\ref{fig:espdata}.
We defer the details of our data and the corresponding collection process to Section~\ref{sec:ax_data_detail} and Section~\ref{sec:ax_data_collection} of the Appendix, respectively.

\begin{table*}[ht]
    \centering
    \begin{tabular}{l|c|c|c|c|c}
        Model & Style & B@4 & M & C & Time {\scriptsize (sec/image)}\\
        \hline
        Pseudo-Align~\cite{laina2019iccv} & $\checkmark$ & 5.2 & 15.5 & 29.4 & - \\ 
        RSA~\cite{honda-etal-2021-removing} & $\checkmark$ & 7.6 & 13.5 & 31.8 & - \\
        Unpaired~\cite{laina2019iccv} & $\checkmark$ & 19.3 & 20.1 & 63.6 & - \\  \hline
        CLIP-Infer~\cite{tewel2022cvpr} &  & 2.6 & 11.5 & 14.6 & 65s \\
        CLIP-Infer-Style & $\checkmark$ & 7.0 & 15.4 & 34.5 & 65s \\
        CLIP-Retrieval & $\checkmark$ & 4.8 & 11.2 & 13.4 & 0.37s \\
        \modelnamewithlogo-Free (GPT-2) & & 6.3 & 13.3 & 29.1 & 0.65s \\  %
        \modelnamewithlogo-Style (GPT-2) & $\checkmark$ & \textbf{21.9} & \textbf{21.9} & \textbf{78.2} & 0.65s \\
    \end{tabular}
    \caption{Unpaired captioning experiments in COCO test split. B@4 denotes Bleu-4, M METEOR and C CIDEr score. Running time entails the whole time for each process needed to infer caption for an image, including image loading and feature extraction. We use greedy decoding for all results in this table.}
    \label{tab:zero_coco_table}
\end{table*}

\begin{table}[ht]
    \centering
    \begin{tabular}{l|c|c|c|c}
        Model & Z\scriptsize{ero-shot} & B@4 & M & C \\
        \hline
        CLIPCap-MLP & & 27.4 & 22.4 & 94.4 \\
        CLIPCap-Full & & 32.2  & 27.1 & 108.4 \\
        \hline
        \modelnamewithlogo-Style & $\checkmark$ & 21.9 & 21.9 & 78.2 \\
        \modelnamewithlogo-MLP & & 31.2 & 25.4 & 103.1 \\
        \modelnamewithlogo-Full & & \textbf{33.1} & \textbf{27.7}  & \textbf{111.1}  \\  %
    \end{tabular}
    \caption{Finetuning experiment in COCO Captions test split.}
    \label{tab:finetune_table}
\end{table}

\begin{table}[ht]
    \centering
    \begin{tabular}{l|c|c|c}
        Model (GPT-2)                  & B@4  & M    & C    \\ \hline
        Audio Prompt + w2c            & 0.17 & 4.03 & 3.14 \\
        Oracle Prompt + w2c           & 0.80 & 5.34 & 7.07 \\ \hline 
        \modelnamewithlogo-Audio-Free         & 0.36 & 3.05 & 4.68 \\
        \modelnamewithlogo-Audio-Style        & \textbf{1.21} & \textbf{6.18} & \textbf{9.54} \\
    \end{tabular}
    \caption{Unpaired audio captioning experiments in AudioCaps test split.}
    \label{tab:audio_table}
\end{table}

\begin{table}[ht]
    \centering
    \setlength{\tabcolsep}{0.3em}
    \begin{tabular}{p{2.5cm}|c|c|c|c}
        \multicolumn{5}{c}{News} \\
        \hline
        Model & Z\scriptsize{ero-shot} & B@4 & M & C \\
        \hline
        \small{Show Attend Tell} & & 0.7 & 4.1 & 12.2 \\
        Text-Only & $\checkmark$ & 0.2 & 2.7 & 1.3 \\
        \hline
        \modelnamewithlogo-Style & $\checkmark$ & 0.8 & 4.4 & 4.6 \\  %
        \modelnamewithlogo-MLP & & \textbf{1.3} & \textbf{4.8} & \textbf{15.7} \\
        \hline
        \multicolumn{5}{c}{Dialog} \\
        \hline
        Model & Z\scriptsize{ero-shot} & NDCG & MRR & R@1 \\
        \hline
        ViLBERT & $\checkmark$ & 11.6 & 6.9 & 2.6 \\
        ViLBERT-Head &  & 19.7 & 9.8 & 3.4 \\
        Text-Only & $\checkmark$ & 19.3 & 18.3 & 5.7 \\
        \hline
        \modelnamewithlogo-Style & $\checkmark$ & \textbf{22.3} & \textbf{25.7} & \textbf{14.6}  \\
    \end{tabular}
    \setlength{\tabcolsep}{1em}
    \caption{Downstream task evaluation in (VisualNews~\cite{liu2021visualnews} test split and VisDial~\cite{das2017visual} validation split. NDCG denotes Normalized Discounted Cumulative Gain, MRR Mean Reciprocal Rank and R@1 Recall at top 1.
    All our results on VisDial are evaluated with the official server.}
    \label{tab:multitask_table}
\end{table}

\paragraph{Training.}
While \modelname could benefit from a more extensive and diverse set of unpaired images,
for fair comparisons with the baselines,
we limit our data to COCO training set images (\emph{unpaired} with their captions). 
We use AdamW~\cite{loshchilov2018decoupled} optimizer ($\beta_2=0.999$, $\epsilon=1e-8$) and fix the learning rate to $1e-5$ with linear decay schedule.
The models are trained until there is no improvement in CLIP cosine similarity for COCO validation set images
up to 50 epochs.
Using a single NVIDIA A6000, and
GPT-2-base/CLIP \texttt{ViT-B/32}
as backbone models, \modelname needs about two days to achieve our reported evaluation scores.

\paragraph{\modelnamewithlogo Models.}
In addition to
\modelname-Free (vanilla GPT-2 as the backbone)
\modelname-Style, we experiment with %
\modelname-MLP, which freezes GPT-2 part of \modelname-Style and finetunes only the light MLP encoder but with supervised MSCOCO (image, caption) pairs and \modelname-Full trains 
the encoder and GPT-2
jointly with supervised MSCOCO (image, caption) pairs. All models use greedy decoding to generate descriptions at inference time.

\subsection{Evaluation of Visual Alignment}
\label{subsec:image_captioning}

We first evaluate the strength of the alignment between an input image and the generated text in \modelname.
First, we %
consider the unsupervised task of unpaired image captioning~\cite{feng2019unpaired}.
Then, we experiment with the usage of the \modelname in task transfer
by comparing the trained weights with randomly initialized ones in a supervised setup.
Following previous works on unpaired captioning~\cite{feng2019unpaired,laina2019iccv},
we split the pairing between image and caption and train them separately using \modelname for unsupervised evaluation.
We split COCO Captions dataset~\cite{lin2014microsoft} with Karpathy split~\cite{karpathy2015deep}.

\subsubsection{Zero-Shot Captioning}
In Table~\ref{tab:zero_coco_table}, we show that \modelname effectively
aligns the image to text without explicitly paired data. Specifically, we compare to the state-of-the-art unpaired captioning methods~\cite{honda-etal-2021-removing,laina2019iccv} and variants of CLIP based decoding methods: CLIP-Infer~\cite{tewel2022cvpr} that uses CLIP to guide GPT2 at inference, CLIP-Infer-Style which runs CLIP-Infer with our style-augmented GPT2 generator and CLIP-Retrieval that retrieves caption with the highest CLIP cosine similarity from the training data.
According to the standard BLEU-4 \cite{papineni2002bleu}, Meteor \cite{banerjee2005meteor}, and CIDEr \cite{Vedantam_2015_CVPR} automatic evaluation metrics,
\modelname achieves superior performance %
against previous
state-of-the-art methods and CLIP based decoding algorithms. %
As stated in previous literature~\cite{feng2019unpaired}, we also reaffirm
that style of the text affects the automatic evaluation to a great deal:
\modelname-Free, which does not know COCO caption text style, falls behind  \modelname-Style (which has been pretrained on unaligned COCO captions, with the prefix \texttt{caption:}).

Finally, note that the computation overhead of \modelname on inference is almost negligible compared to that of CLIP-Infer, a decoding time method~\cite{tewel2022cvpr}.
On inference time, \modelname's runtime is comparable to vanilla GPT-2 alone.
Only the lightweight encoder needs to run on top of GPT-2,
offering fast inference speed.

\subsubsection{Finetuning}

As our policy network shares the same architecture with MLP-variant CLIPCap~\cite{mokady2021clipcap},
we can directly evaluate the contribution of our encoder
as pretrained weights in a supervised setting.
In Table~\ref{tab:finetune_table}, we show \modelname initialization bests random initialization both when updating and fixing GPT parameters.
Thus, our framework can provide efficient initial alignment between two pretrained modules.

\subsection{Evaluation of Auditory Alignment}
\label{subsec:audio_captioning}

We extend \modelname to another modality: audio.
As an auditory counterpart of CLIP, we use Wav2CLIP~\cite{wu2022wav2clip} to score the audio-linguistic alignment during RL training, but otherwise, the setup remains the same.
Here, we break the pairing in an audio captioning dataset AudioCaps~\cite{kim2019audiocaps} to evaluate unpaired audio captioning performance.
We follow an identical evaluation protocol as in \S~\ref{subsec:image_captioning}, except that we only use audio as input.

In Table~\ref{tab:audio_table}, we only report the performance of GPT-2-based baselines, as the unpaired image baselines \cite{laina2019iccv,honda-etal-2021-removing} require object detectors and cannot be directly applied. %
\modelname achieves better results than baseline models,
which first rollout random text samples conditioned on fixed (\eg \texttt{Sound of a}) or the oracle prompts and then select ones with maximal CLIP cosine similarity.
Also, the style prompt tuning positively contributes to \modelname`s performance, increasing CIDEr by 4.86.
Wav2CLIP (and preliminary experiments with other audio encoders, specifically, \newcite{guzhov2022audioclip,wu2022wav2clip} which are also pretrained on an audio classification dataset~\cite{gemmeke2017audioset,chen2020vggsound}) appears to provide less accurate training signal for \modelname
compared to image CLIP pretrained on large-scale image caption dataset~\cite{radford2021learning}.
We expect this is the case not only because audio classification datasets are relatively small~\cite{zhao2021connecting} %
 but also because these datasets do not offer rich natural language annotations.
Still, our model can generate audio-relevant and plausible captions as described in Figure~\ref{fig:qual_audio}.

\subsection{Generalization to Diverse Styles}
\label{subsec:multitask}

We now experiment beyond standard image captioning setups
to demonstrate \modelname's capacity to generate image-related texts of diverse styles.
Here, we evaluate two styles that can be supported by existing public corpora: visual news and dialogue.

\begin{figure*}[t]
\centering
\includegraphics[trim=0.0cm 0.0cm 0cm 0.0cm,clip,width=\textwidth]{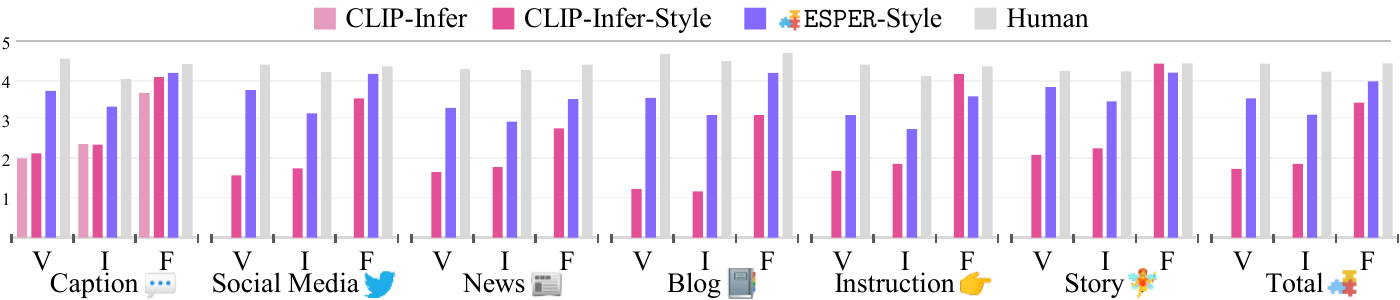}
\caption{Human evaluation of captions for each style prompt. We take the average of 5-point Likert-scale rating from three annotators. V denotes visual relevance, I informativeness and F for fluency. }
\label{fig:humaneval}
\end{figure*}
\begin{figure}[t]
\centering
\includegraphics[trim=0.0cm 0.0cm 0cm 0.0cm,clip,width=0.48\textwidth]{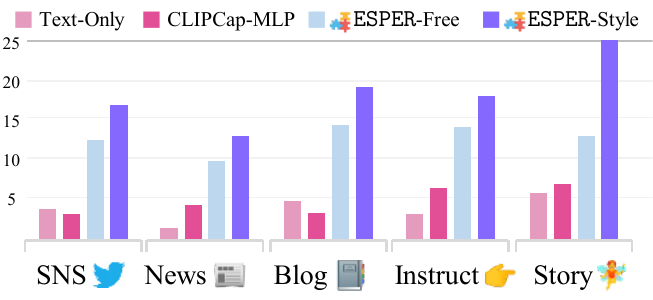}
\caption{Evaluation on the ESP dataset. We report CIDEr in this plot.}
\label{fig:espeval}
\end{figure}

\subsubsection{Visual News}
VisualNews~\cite{liu2021visualnews} includes 1.08 million news images along with associated image captions and articles, sourced from four news sites.
The captions describe the image's relevance to the news article instead of simply describing the literal image contents.
For our experiments, we assign respective style prefixes per news source.
For a fair comparison, we compare \modelname with models that rely only on image inputs,\footnote{
Other baselines for VisualNews generate based on the article text or keywords as inputs
and hence are not directly comparable to our framework.} e.g., Show Attend Tell~\cite{xu2015show}, from \newcite{liu2021visualnews}.
We also include the text-only style generator without visual inputs as another baseline (Text-Only). %

Results are in Table~\ref{tab:multitask_table}:
zero-shot \modelname outperforms not only the text-only baseline but also
the supervised baseline in Bleu-4 and METEOR scores.
However, it lags behind the supervised model by a wide margin in CIDEr terms of CIDEr.
We attribute this difference to a combined effect of the news style and CLIP:
while news consists of a myriad of proper nouns,
CLIP has not been exposed to a majority of such terms.
As a result, \modelname does not generate as many proper nouns as in the ground truth captions, decreasing the CIDEr score, which takes the rarity of terms into account.
By finetuning the adaptor, \modelname overcomes this knowledge gap
and surpasses the baselines even in the CIDEr score.
\subsubsection{Visual Dialogue}

VisDial~\cite{das2017visual} is a dataset of iterative dialogues conditioned on an image.
Given an image and previous dialogue act,
the model is asked to rank the likelihood of the 100 answer candidates.
After training \modelname with the unpaired dialogue style generator backbone, %
we rank the answer candidates by likelihood of the answers given the image and the question.
We use the validation set for evaluation for fair comparison against previously reported zero-shot baseline results~\cite{murahari2020visdialbert}.
The baselines consist of ViLBERT~\cite{lu2019vilbert} and frozen ViLBERT~\cite{lu2019vilbert} finetuned with a linear head.

The bottom half of Table~\ref{tab:multitask_table} shows the VisDial dataset re-ranking results.
Zero-shot \modelname improves the baselines by a margin. It even outperforms the supervised ViLBERT-Head,
showing that \modelname is capable of discerning likely visual dialogues.

\subsection{From One Image to Many Styles}
\label{subsec:exp_our_dataset}

While we observe that \modelname can generate diverse image-related texts,
we still need to prove that this diversity in style is induced by text prompts.
A null hypothesis is that there are identifiable and consistent features found, e.g., only in news articles.
The model may have exploited this superficial relation to generating news style captions.
\datasetname from Section~\ref{sec:data} is specifically designed to counter this hypothesis as it exhibits \emph{multiple styled texts} for \emph{the same image.}

Figure~\ref{fig:espeval} we show that 
\modelname can generate diverse text depending on textual style prompts.
\modelname outperforms CLIPCap-MLP~\cite{mokady2021clipcap}, a COCO-supervised baseline,
demonstrating prompt-conditioned generation is necessary to handle \datasetname.
Also, the text-only baseline falls by a wide margin,
indicating that the visual-linguistic alignment is as important as the text diversity.
Finally, \modelname-Style improves over \modelname-Free
to show the effect of explicit style conditioning.
For fine-grained results, refer to Table~\ref{tab:esp_large} 
in Appendix~\ref{sec:ax_esp_experiment}.

\subsection{Human Evaluations on \datasetname}
\label{subsec:humaneval}
We conduct a human evaluation on \modelname, CLIP-Infer\footnote{We use a version of GPT2-base size model to generate descriptions to be comparable to our generation framework.}, and CLIP-Infer-Style generated descriptions as well as ground truth captions 
that complete the following six prompts (\texttt{caption:}, \texttt{social media:}, \texttt{news:}, \texttt{blog:},  \texttt{instruction:}, \texttt{story:})
We choose random 100 images in \datasetname test split and ask English-proficient human annotators to provide a 5-point Likert-scale if the sentences: 1) are visually relevant to the image (Vis), 2) provide informative and interesting content for the prompt (Inf), 3) and sound fluent and human-like (Flu). Each sample is evaluated by three annotators using the Amazon Mechanical Turk platform. The results are shown in Figure~\ref{fig:humaneval}. %
On average, \modelname provides more visually relevant and informative content in every prompt than CLIP-Infer. While CLIP-Infer has slightly more fluent (Flu) descriptions on the story and instruction prompt, we found that this is likely due to their descriptions being relatively short, thus having less room for grammatical errors. %

\begin{figure}[t]
\centering
\includegraphics[trim=0.0cm 0.0cm 0cm 0.0cm,clip,width=0.5\textwidth]{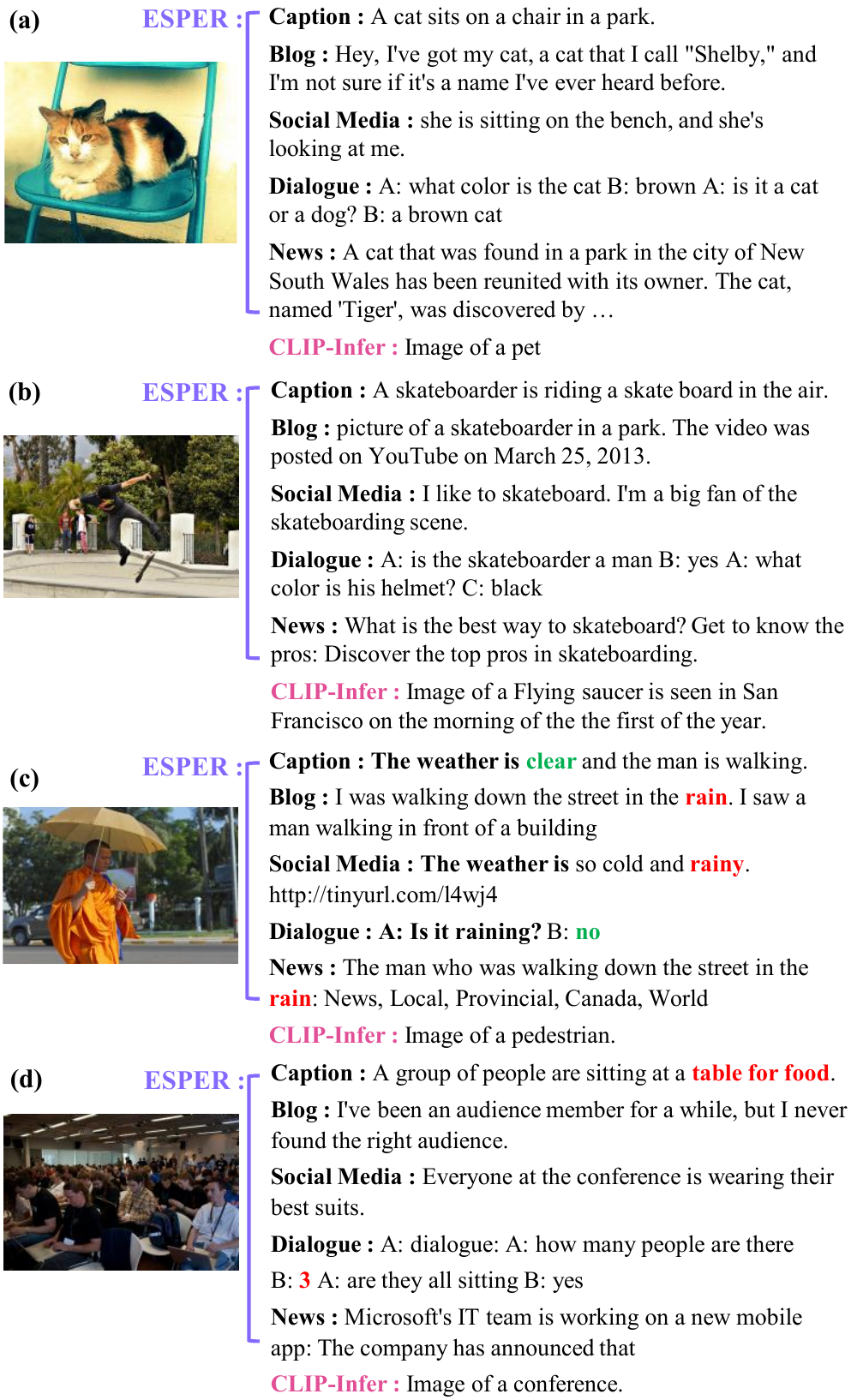}
\caption{\modelnamewithlogo Zero-shot image captioning examples on various style prompts.
The conditioning text prompt is denoted in bold(\ie \textbf{``text''}).
We mark visually relevant points with \textbf{\textcolor{figure_green}{green}} and errors with \textbf{\textcolor{red}{red}}.
}

\label{fig:qual_style}
\end{figure}

\begin{figure}[t]
\centering
\includegraphics[trim=0.0cm 0.0cm 0cm 0.0cm,clip,width=0.5\textwidth]{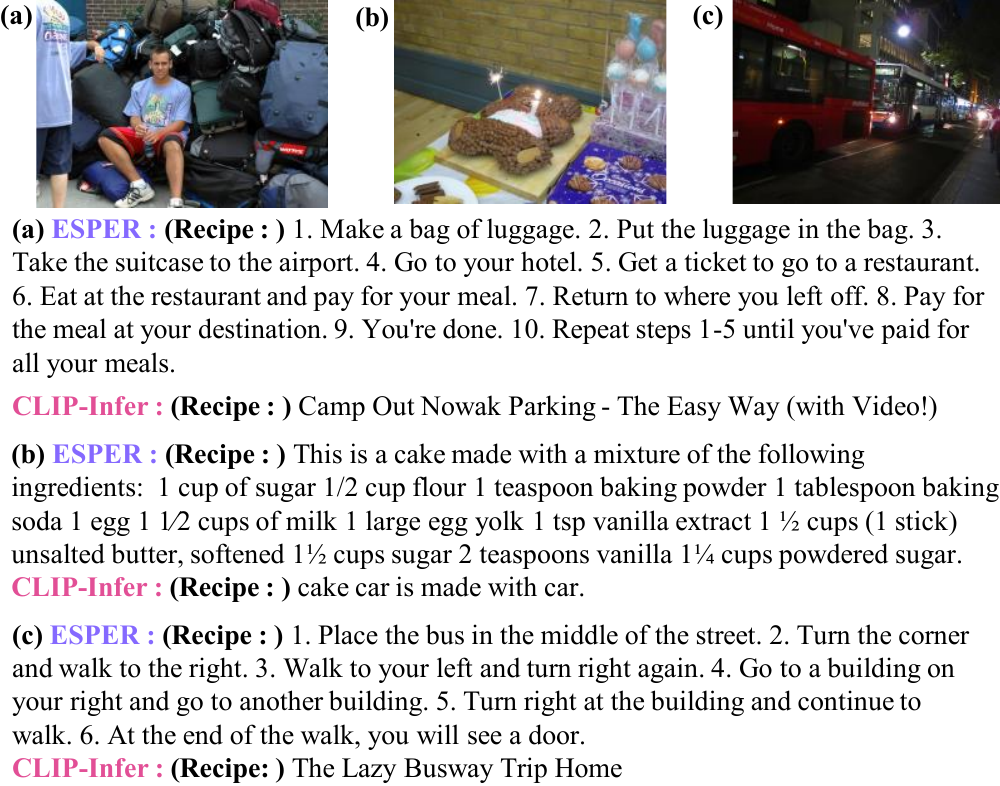}
\caption{\modelnamewithlogo generation results on custom task prompts, \textbf{(Recipe: )}. \modelname has never trained on recipe prompts. }
\label{fig:qual_custom}
\end{figure}

\begin{figure}[t]
\centering
\includegraphics[trim=0.0cm 0.1cm 0cm 0.0cm,clip,width=0.5\textwidth]{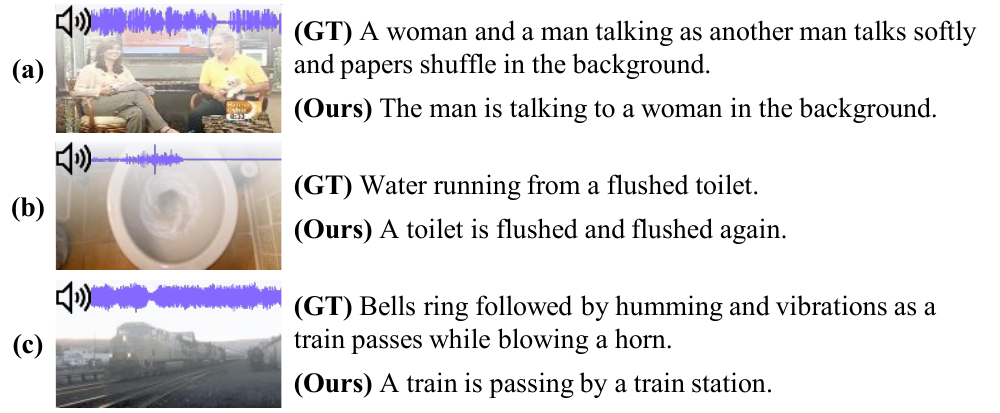}
\caption{\modelnamewithlogo generation results on zero-shot audio captioning.
Each image is the keyframe of the original video for illustration purposes. 
\modelname-Audio uses only audio without visual input.
}

\label{fig:qual_audio}
\end{figure}

\subsection{Qualitative Results}
\label{subsec:qualitative}

Figure ~\ref{fig:qual_style},~\ref{fig:qual_custom} presents zero-shot captioning results on COCO images generated by \modelname-style and CLIP-Infer baseline~\cite{tewel2022cvpr}\footnote{We used their public demo for qualitative results. https://replicate.com/yoadtew/zero-shot-image-to-text}. 

Figure~\ref{fig:qual_style} shows some diverse zero-shot captions from COCO test split.
Conditioning on both image and prefix, \modelname generates various visually sensible and informative captions.
But Fig~\ref{fig:qual_style}.(c) and (d) show inaccurate caption compare to the CLIP-Infer baseline. In (c), while the monk is holding an umbrella,  we can deduce that it is not raining from the clear sky. 
However, \modelname confuses the weather condition depending on the text prompt. Also, the model suffers from false bias in visual counting. \textit{((d) Dialogue A: how many people are there B: 3)}

Figure~\ref{fig:qual_custom} shows generation results on the ``recipe" task prompt that was \emph{not previously pre-trained as a style prompt.}
\modelname generate not only sensible cake recipe generation in Fig~\ref{fig:qual_custom}.(b), but also reasonable ``recipe" even when it is not conditioned on a food image (Fig~\ref{fig:qual_custom}.(a),(c)); similar performance was observed for ``My favorite poem" and ``lyrics" that GPT-2 can generate. In most cases, CLIP-Infer generally produces short generations; and, because it wasn't designed to adapt to individual styles, it cannot as effectively generalize to custom prompts. Figure~\ref{fig:qual_audio} additionally demonstrates that \modelname can also adapt to the audio via wav2clip rewards.

\section{Related Work}
\label{sec:related_work}

\textbf{Visual-Language Pretraining}.
Successful vision-language models pretrained on large-scale image-text corpora have been proposed, e.g., BERT-style~\cite{Devlin2019BERTPO} models \citet{Tan2019LXMERTLC, Chen2020UNITERUI, Li2020OscarOA,zellers_2021_merlot}, encoder-decoder style models, \citet{Zhou2020UnifiedVP,Wang2021SimVLMSV,Jia2021ScalingUV}
and constrastive models \cite{radford2021learning,jia2021scaling}.
Vision-text models have additionally been extended to audio \cite{zhao2021connecting,zellers2022merlot}.
TAPM~\cite{Yu_2021_CVPR} adapts visual encoder and GPT with self-supervised training objective that predicts causal order of the visual story.

\textbf{Multimodal prompt tuning}. 
Prefix tuning~\cite{li2021prefix} and Prompt tuning~\cite{lester2021power} %
simplify finetuning large models %
by all but a small number of parameters. %
\newcite{tsimpoukelli2021multimodal} adapt prefix tuning to images via maximum likelihood training a small image-to-text adapter using Conceptual Captions~\cite{sharma2018conceptual}.
Like \modelname, CLIPCap~\cite{mokady2021clipcap} %
combines GPT + CLIP image features to generate image caption.
We use the same architecture as in CLIPCap and fix GPT weights likewise, effectively following the setup of p-tuning~\cite{liu2021gpt}.

\paragraph{Unsupervised captioning.}
To learn visual-linguistic relationship without paired data, previous literature draws upon 
pseudo-pairing retrieved with visual concept detector~\cite{honda-etal-2021-removing} or joint image-language embedding space~\cite{laina2019iccv}.
Most related to our work is \citet{tewel2022cvpr} that uses CLIP image-text alignment score to guide inference of pretrained language model without further training.

\paragraph{Reinforcement learning for language tasks.}
In image captioning, RL has been used to resolve the discrepancy between training and inference data~\cite{ranzato2016sequence,bengio2015scheduled}
or to optimize discrete language metrics directly~\cite{rennie2017self}.
Storytelling models employ RL to maintain coherence in the story~\cite{tambwekar2018controllable} or incorporate human feedback~\cite{Martin2017a}.
RL is also proven effective in goal-driven dialogue~\cite{ammanabrolu2022dialogue}, interactive QA~\cite{yuan-etal-2019-interactive}, grounded generation in text games~\cite{hausknecht19,wang2022scienceworld} and value alignment to human preferences~\cite{nahian19norm,hendrycks2021jiminycricket,ammanabrolu2022aligning}.
Recently, Instruction GPT~\cite{ouyang2022instructiongpt} shows RL can improve prompt-conditioned generation quality of pretrained language models.
To the best of our knowledge, \modelname is the first method to use multimodal reward to align images to pretrained language models; while
\citet{cho2022clipreward} used CLIP rewards as well they finetune an already trained image captioning model
instead of a general large language model.

\section{Conclusion}
\modelnamewithlogo combines language generation capability in GPT-2 with multimodal knowledge in CLIP to build a diverse image-conditioned text generator: instead of maximum likelihood training, we train via reinforcement learning rewards.
We note that the RL objective we consider can be used in conjunction with multimodal prompt tuning~\cite{tsimpoukelli2021multimodal} and zero-shot captioning with CLIP guidance~\cite{tewel2022cvpr}. %

Future work includes: 
\begin{enumerate}[leftmargin=*,topsep=0pt,itemsep=-1ex,partopsep=1ex,parsep=1ex]
\item enhancing \modelname so that it can simultaneously maximize rewards for multiple modalities (Image, Audio, OCR, Motion in video, etc.);
\item scaling up the CLIP and GPT-2 backbones to larger variants; and
\item exploring the utility of \modelname as a data augmentation tool for multimodal reasoning tasks. %
\end{enumerate}

\section{Acknowledgements}
We express special thanks to the Mosaic team members and AI2 researchers who gave feedback on this project.
Also, we express our gratitude for the helpful comments 
by Jaekyeom Kim on reinforcement learning techniques.
This work was supported by the Allen Institute for AI and 
DARPA MCS program through NIWC Pacific (N66001-19-2-4031).
SNU members are supported by Basic
Science Research Program through the National Research
Foundation of Korea (NRF) (2020R1A2B5B03095585)
and Institute of Information \& communications Technology
Planning \& Evaluation (IITP) grant (No.2019-0-01082, SW
StarLab). 
We thank all our workers on MTurk for their contributions
to our project.

\bibliography{main}
\bibliographystyle{acl_natbib}

\appendix

\clearpage
\section{Language Model Backbones}
\label{sec:ax_language_model_backbones}
\begin{figure}[h]
\centering
\includegraphics[trim=0.0cm 0.0cm 0cm 0.0cm,clip,width=0.5\textwidth]{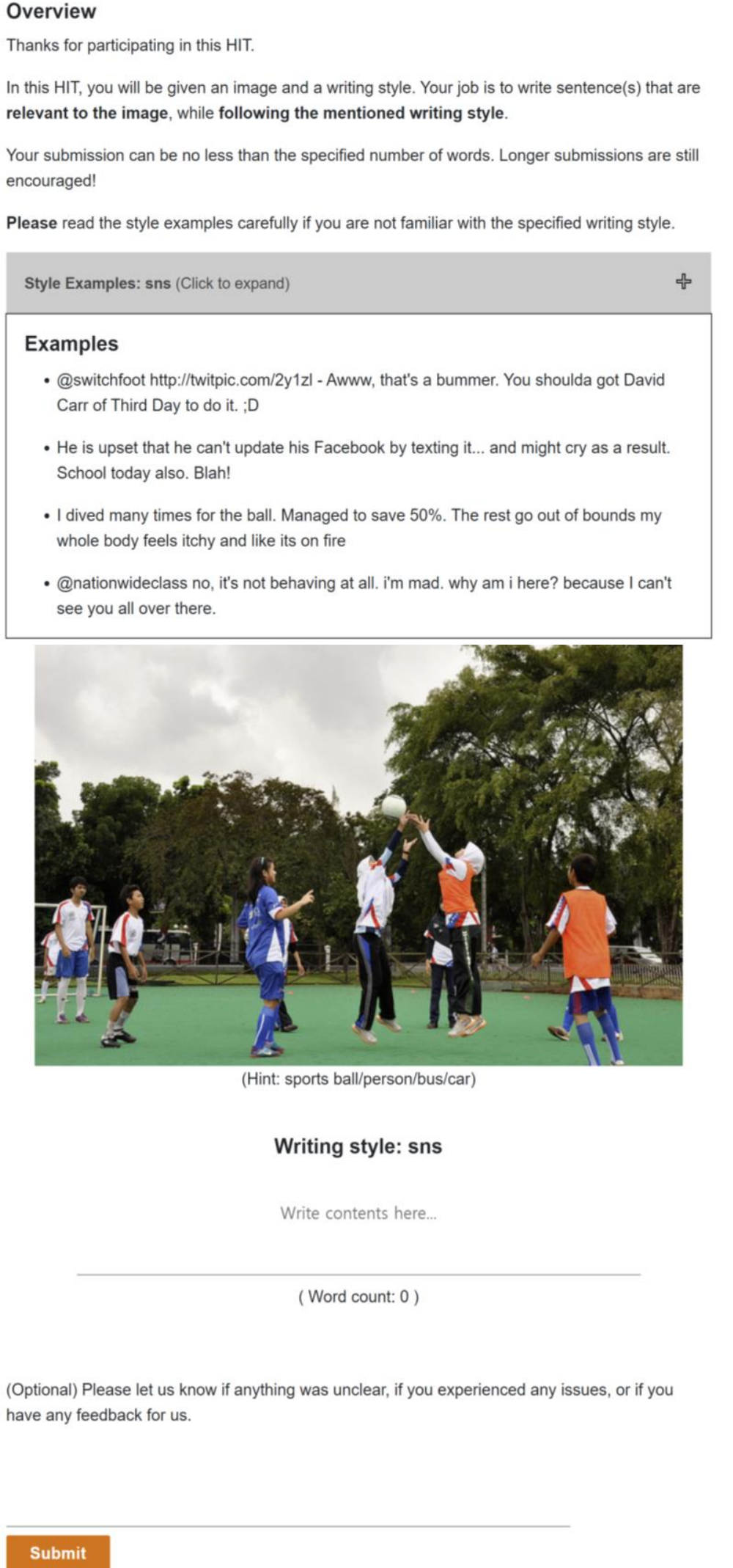}
\caption{Annotation interface of \datasetname.}

\label{fig:dataset_ui}
\end{figure}
\paragraph{\modelname-Free.}
We use GPT-2-base~\cite{radford2019language} as the language model backbone for all experiments.
Since GPT-2 does not have a special start-of-sentence token,
we provide a random single token as an initial text prompt to start generation on.
This initial token is sampled from GPT-2 vocabulary with sampling weight computed with token frequency.

\paragraph{\modelname-Style.}
As summarized in Section~\ref{sec:style_ptuning}, we finetune GPT-2
on text-only corpus with style prompts to prepare the style generator backbones.
The style prompts and corresponding text corpus sources include:

\begin{itemize}
    \item \texttt{caption:} COCO Caption~\cite{lin2014microsoft}
    \item \texttt{social media:} \begin{itemize}
        \item Sentiment140~\cite{go2009twitter}
        \item MDID~\cite{kruk2019integrating}
        \item TweetEval~\cite{barbieri2020tweeteval}
    \end{itemize}
    \item \texttt{news:} GoodNews~\cite{biten2019good}
    \item \texttt{blog:} Blog Authorship~\cite{schler2006effects}
    \item \texttt{instruction:} WikiHow~\cite{koupaee2018wikihow}
    \item \texttt{story:} \begin{itemize}
        \item ROCStories~\cite{mostafazadeh2016corpus}
        \item TimeTravel~\cite{qin2019counterfactual}
    \end{itemize}
\end{itemize}

For visual news generation, we use different style prompt per news source
(\texttt{bbc:}, \texttt{guardian:}, \texttt{usa today:}, \texttt{washington post:})
to reflect writing style differences between media in VisualNews dataset~\cite{liu2021visualnews}.

\section{Language Model RL Training}
\label{sec:ax_stability_rewards}

\begin{table*}[ht]
    \centering
    \small
    \addtolength{\tabcolsep}{-3pt}    
    \begin{tabular}{l|c|ccc|ccc|ccc|ccc|ccc|ccc}
        && \multicolumn{3}{c}{Social Media}
        & \multicolumn{3}{c}{News}
        & \multicolumn{3}{c}{Blog}
        & \multicolumn{3}{c}{Instruction}
        & \multicolumn{3}{c}{Story}
        & \multicolumn{3}{c}{Total} \\ \hline
        Model         
        & Prompt 
        & B & M & C
        & B & M & C
        & B & M & C
        & B & M & C
        & B & M & C
        & B & M & C \\ \hline
        Text-Only 
        & $\checkmark$
        & 0.2& 3.7& 3.9& 0.0& 2.2& 1.6& 0.3& 4.1& 4.9& 0.0& 4.0& 3.3& 0.3& 4.7& 5.9& 0.1& 3.7& 3.9 \\ \hline
        \multirow{2}{*}{ClipCap-MLP}
        &
        & 0.0& 3.9& 6.8& 0.0& 4.8& 7.5& 0.3& 4.0& 6.6& 0.3& 4.2& 7.6& 0.0& 4.3& 7.3& 0.1& 4.2& 7.2 \\ %
        & $\checkmark$
        & 0.2& 3.0& 3.3& 0.2& 3.9& 4.5& 0.0& 2.9& 3.4& 0.5& 4.8& 6.5& 0.0& 4.4& 7.1& 0.2& 3.8& 5.0 \\  %
        \hline
        \modelname-Free
        & $\checkmark$
        & \textbf{0.6}& 5.6& 12.5& 0.6& 5.5& 9.9& \textbf{0.7}& 6.2& 14.4& \textbf{0.7}& 5.6& 14.1& 0.6& 5.7& 13.0& 0.6& 5.7& 12.8 \\
        \modelname-Style 
        & $\checkmark$
        & \textbf{0.6}& \textbf{5.8}& \textbf{16.9}& \textbf{0.7}& \textbf{5.7}& \textbf{13.0}& \textbf{0.7}& \textbf{6.7}& \textbf{19.2}& \textbf{0.7}& \textbf{5.7}& \textbf{18.0}& \textbf{1.2}& \textbf{7.5}& \textbf{25.0}& \textbf{0.8}& \textbf{6.3}& \textbf{18.4} \\ \hline
    \end{tabular}
    \addtolength{\tabcolsep}{+3pt}    
    \caption{Style-wise experiment results on \datasetname. B denotes Bleu-4 score.}
    \label{tab:esp_large}
\end{table*}
\begin{table*}[ht]
    \centering
    \small
    \scalebox{0.88}{
    \addtolength{\tabcolsep}{-3pt}
    
    \begin{tabular}{l|ccc|ccc|ccc|ccc|ccc|ccc|ccc}
        & \multicolumn{3}{c}{Caption}
        & \multicolumn{3}{c}{Social Media}
        & \multicolumn{3}{c}{News}
        & \multicolumn{3}{c}{Blog}
        & \multicolumn{3}{c}{Instruction}
        & \multicolumn{3}{c}{Story}
        & \multicolumn{3}{c}{Total} \\ \hline
        Model
        & Vis. & Inf. & Flu.
        & Vis. & Inf. & Flu.
        & Vis. & Inf. & Flu.
        & Vis. & Inf. & Flu.
        & Vis. & Inf. & Flu.
        & Vis. & Inf. & Flu.
        & Vis. & Inf. & Flu.\\
        \hline
        Clip-Infer
        & 1.98 & 2.34 & 3.62 
        &  -  &   -   &  -
        &  -  &   -   &  -
        &  -  &   -   &  -
        &  -  &   -   &  -
        &  -  &   -   &  -
        &  -  &   -   &  -\\
        Clip-Infer-Style
        & 2.11 & 2.33 & 4.01
        & 1.56 & 1.73 & 3.48 
        & 1.64 & 1.76 & 2.73
        & 1.21 & 1.16 & 3.06
        & 1.67 & 1.85 & \textbf{4.09}
        & 2.07 & 2.23 & \textbf{4.35}
        & 1.72 & 1.85 & 3.38 
        \\
        \modelname-Style
        & \textbf{3.67} & \textbf{3.27} & \textbf{4.12}
        & \textbf{3.69} & \textbf{3.11} & \textbf{4.10}
        & \textbf{3.24} & \textbf{2.90} & \textbf{3.46}
        & \textbf{3.49} & \textbf{3.06} & \textbf{4.12}
        & \textbf{3.06} & \textbf{2.71} & 3.53
        & \textbf{3.76} & \textbf{3.41} & 4.13
        & \textbf{3.48} & \textbf{3.08} & \textbf{3.91}
        \\
        \hline
        Human 
        & 4.47 & 3.96 & 4.34
        & 4.32 & 4.14 & 4.28
        & 4.21 & 4.19 & 4.33
        & 4.60 & 4.41 & 4.62
        & 4.32 & 4.04 & 4.28
        & 4.17 & 4.16 & 4.36
        & 4.35 & 4.15 & 4.36
        \\
        \bottomrule
    \end{tabular}
    
    \addtolength{\tabcolsep}{+3pt}
    }
    \caption{Human evaluation of captions for each style prompt. We take the average of 5-point Likert-scale rating from three annotators. Vis. denotes visual relevance, Inf. informativeness and Flu. for fluency.}
    \label{tab:human_eval}
\end{table*}

\paragraph{KL Divergence.}
By constraining KL divergence between
the online policy and the initial language model,
we aim to maintain salience of the generated text.
Here, we simply optimize the difference between
the log likelihood of the online policy and the initial policy
for each token generated.

\paragraph{Reference Entropy.}
To constrain deviation from text generation capability,
we first compute text-only log likelihood using
either the pretrained text style generator or the vanilla language model.
Then, we penalize the model whenever the text-only negative log likelihood of a generated token exceeds a predefined threshold $\tau_e = \frac{70}{l}$, where $l$ is the length of the generated sequence.
We take inverse of the difference between negative log likelihood and threshold and optimize it as a reward.
In practice, we further scale this reward with fixed gain $\alpha_e = 0.1$.

\paragraph{Repetition Penalty.}
This reward penalizes the model for generating repeated n-grams.
Given GPT tokenizer, we count repeated (1, 2, 3)-grams.
Specifically, we subtract the number unique of n-grams from 
that of all n-grams to count repetitions.
Then we compute a weighted sum of the n-gram repetition counts
and scale the combined score with fixed gain
$\alpha_r = 0.025 $ and bias $\beta_r = 0$.

\section{Details on \datasetname}
\label{sec:ax_data_detail}

There are multiple ways to describe an image depending on the context and intent of the author. We refer to these multiple methods as \emph{styles}.
Previous works focus on the sentiment of a caption like positive \& negative~\cite{mathews2016senticap}, romantic \& humorous~\cite{gan2017stylenet},
and various personalities~\cite{shuster2019engaging}. 
However, style does not solely depend on sentiments and emotions: 
it comprises every choice of text type, structure and vocabulary  used to convey intended meaning of the writer.
As intention of a writer depends on where the one's interest lies,
different information of the same visual cue would be illustrated on each style of writing.

For example, consider an image of a boy with a bow tie singing as part of a choir on a stage.
While this image may have been uploaded by the singer's sibling with a caption like ``go bro, love the bowtie!", a local news article about the same concert might instead write: ``the choir's performance on August 17th went off without a hitch."
Because different styles of writing may focus on different aspects of an image,
and styles may not be fully translatable via text-only operators such as text style transfer, e.g., ``go bro, love the bowtie!" doesn't mention anything about a choir performance.

We thus collect \datasetname to explore broad range of text styles conditioned on the same image.
Using Amazon Mechanical Turk, we ask the annotators to write captions relevant to an image while following writing styles mentioned above.
An image cost about $\$0.3$ to annotate, which translates to $\$7\mathhyphen{}28$ of payment per a work hour depending on the proficiency of the worker.
The average length of \datasetname is 28.4 words (2.3 sentences), and the collected captions are filtered with respect to their adherence to given images and styles.

\section{\datasetname Collection Process}
\label{sec:ax_data_collection}

We use Amazon Mechanical Turk to collect captions as shown in Figure~\ref{fig:dataset_ui}.
For images in COCO Captions test set with respect to Karpathy split, we randomly select images with one to five annotated objects to select images with salient but not noisy context.
We ask the annotators to write sentences that are relevant to the image while following the mentioned writing style.
We provide examples from well-constructed datasets as references, as listed in text corpus sources of Appendix~\ref{sec:ax_language_model_backbones}.
We ask the annotators to write no less than 30 words, but for writing styles with shorter text like social media and news, we lower the bar from 30 to 10 words.
We also regularly monitor the collection so that only the workers with fluency and understanding of style can participate in the process.
In total, 189 workers participated in the collection process.
The collected dataset is filtered by manually verifying whether the captions are relevant to given images and styles.

\section{\datasetname Experiment Details}
\label{sec:ax_esp_experiment}

We compare \modelname against three baselines in this experiment.
The first is a text-only baseline.
We use the pretrained text style generator with random sampling to generate the candidate texts.
The rest two baselines~\cite{mokady2021clipcap} are trained on a caption supervision dataset (COCO captions) and share the same architecture as our \modelname.
As the supervised baselines are not intended for prompt conditioning,
we report evaluation results both with and without the style prompts for them.
When not using the style prompts, we fix the prompt to "Image of a",
following the recommended approach in literature~\cite{mokady2021clipcap}.
For fair comparison against the baselines trained with
supervised dataset of limited length (ClipCap-MLP),
we truncate all text including the ground truth captions
to the first 20 byte-pair tokens with GPT tokenizer.
Note that all compared methods share the same tokenization scheme as the vanilla GPT2
and hence the truncation does not favour any specific approach.

We report the evaluation results in Table~\ref{tab:esp_large}.
For clarification, the scores in Table~\ref{tab:esp_large} include and expand upon
the summarized results in Figure~\ref{fig:espeval}.
\modelname shows flexible adaptability to each style without being exposed to any paired image-text data of the given styles.
On the other hand, the supervised baselines exhibit limited generalizability to diverse text styles even when conditioned on style prompts.
The total score is computed as the mean over metrics of each style,
without considering sample size difference.

\begin{figure}[t]
\centering
\includegraphics[trim=0.0cm 1.0cm 0cm 0.0cm,clip,width=0.5\textwidth]{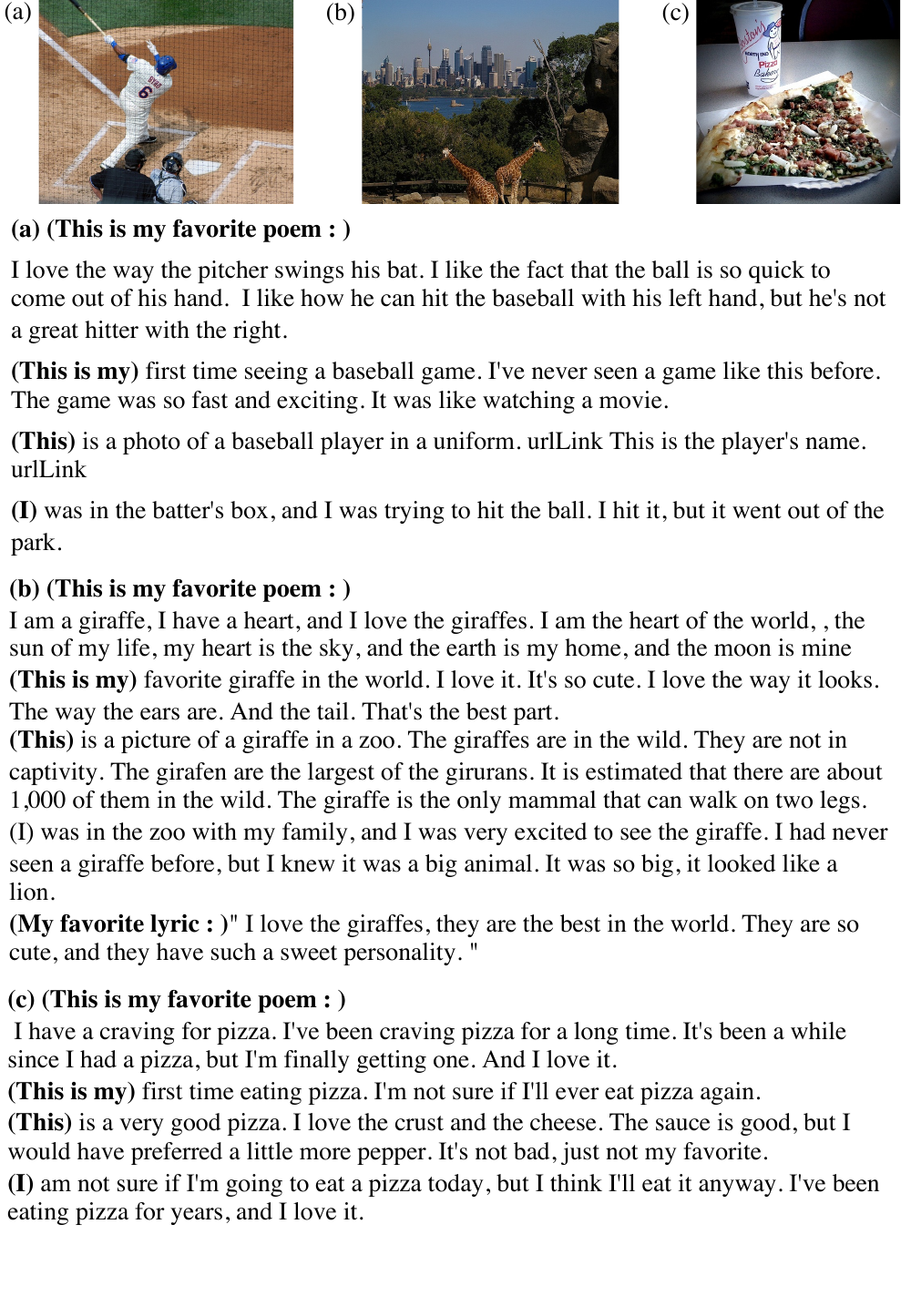}
\caption{\modelname generation results conditioned on custom prompts. 
    }

\label{fig:supp_qual1}
\end{figure}

\begin{figure}[t]
\centering
\includegraphics[trim=0.0cm 1.0cm 0cm 0.0cm,clip,width=0.5\textwidth]{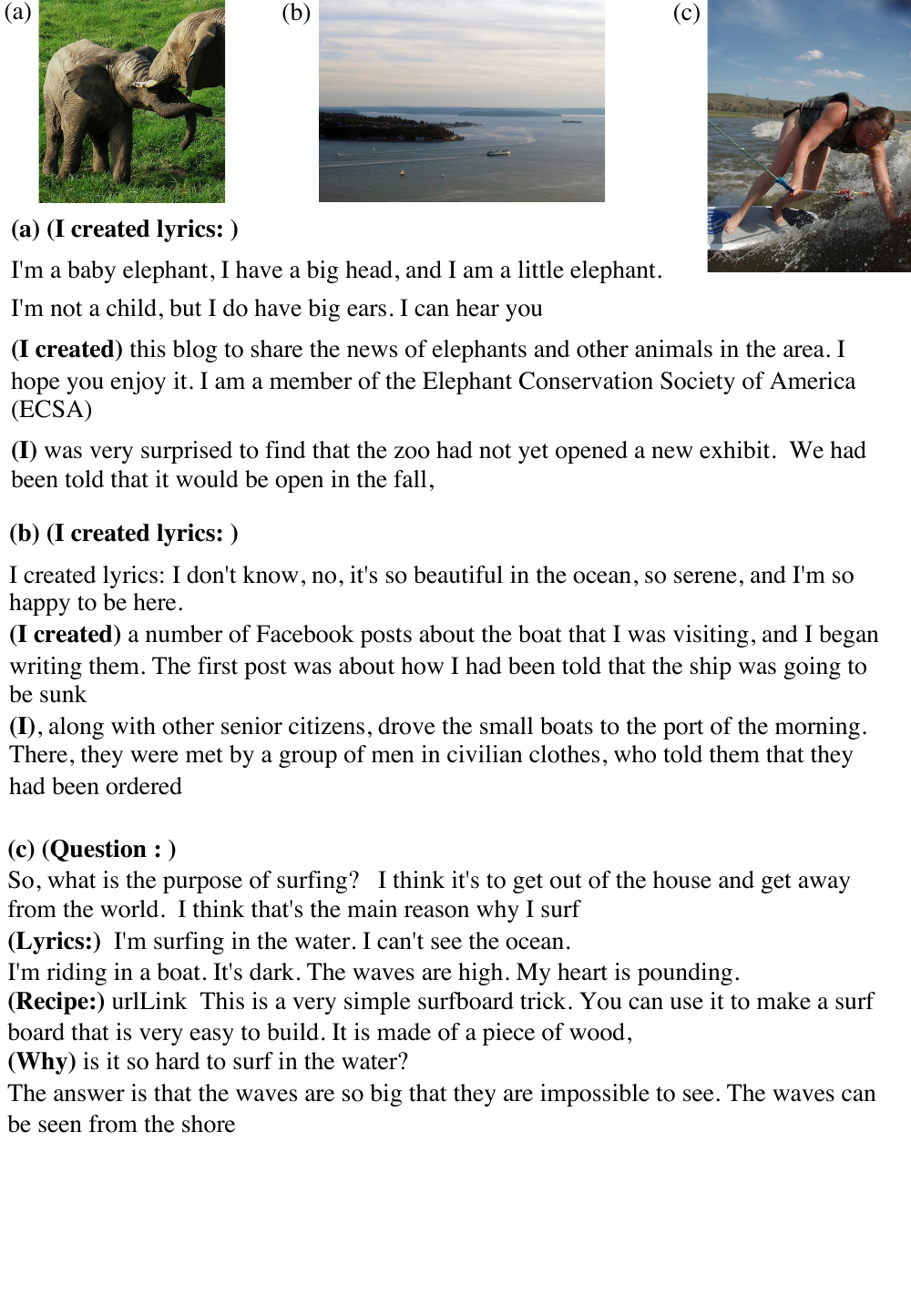}
\caption{\modelname generation results conditioned on custom prompts.}

\label{fig:supp_qual2}
\end{figure}

\section{Additional Qualitative Samples}
\label{sec:ax_qualitative_samples}

In Figure~\ref{fig:supp_qual1}-~\ref{fig:supp_qual2}, we displays \modelname generation results conditioned on custom prompts such as (\textit{This is my favorit poem}) or (\textit{I created lyrics}). The conditioning text prompt is denoted as bold font enclosed with parenthesis (\ie \textbf{"(text)"}). To qualitatively emphasize the randomness of our results, we provided the model with progressively growing prompts.

\end{document}